\documentclass[journal]{IEEEtran}
\usepackage{times}
\usepackage{epsfig}
\usepackage{graphicx}
\usepackage{amsmath}
\usepackage{amssymb}

\usepackage{multirow}
\usepackage[pagebackref=true,breaklinks=true,colorlinks,bookmarks=false]{hyperref}
\newcommand{\etal}{\textit{et al.}}

\usepackage{ifpdf}
\usepackage{cite}
\usepackage{amsmath}
\usepackage{stfloats}

\ifCLASSOPTIONcaptionsoff
 \usepackage[nomarkers]{endfloat}
\let\MYoriglatexcaption\caption
\renewcommand{\caption}[2][\relax]{\MYoriglatexcaption[#2]{#2}}
\fi
\usepackage{url}

\usepackage[normalem]{ulem}
\usepackage{subcaption}

\begin{document}
% ----------------- Response Letter ----------------------------
%\input{response_letter.tex}
% ------------------ Revised Manuscript ----------------------
\title{Adaptive Graph Representation Learning for Video Person Re-identification}

\author{Yiming~Wu,
        Omar El Farouk Bourahla,
        Xi~Li*,
        Fei~Wu,
        and~Qi~Tian,~\IEEEmembership{Fellow,~IEEE},
        Xue~Zhou % <-this % stops a space
\thanks{This work is in part supported by key scientific technological innovation research project by Ministry of Education, Zhejiang Provincial Natural Science Foundation of China under Grant LR19F020004, the National Natural Science Foundation of China under Grants (61751209, 6162510, and 61972071), Baidu AI Frontier Technology Joint Research Program, Zhejiang University K.P.Chao's High Technology Development Foundation, and Zhejiang Lab (No.2019KD0AB02).}
\thanks{Y. Wu, Omar, F. Wu are with College of Computer Science, Zhejiang University, Hangzhou 310027, China (e-mail: ymw, obourahla@zju.edu.cn wufei@cs.zju.edu.cn).}
\thanks{X. Li*(corresponding author) is with the College of Computer Science and Technology, Zhejiang University, Hangzhou 310027, China (e-mail: xilizju@zju.edu.cn).}
\thanks{Q. Tian is with the Department of Computer Science, University of Texas, San Antonio, TX 78249-1604 USA (e-mail: qitian@cs.utsa.edu).}
\thanks{X. Zhou is with the school of automation engineering, University of Electronic Science and Technology of China (e-mail: zhouxue@uestc.edu.cn).}
}

\markboth{IEEE TRANSACTIONS ON IMAGE PROCESSING, VOL. XX, NO. X, 201X}%
{Shell \MakeLowercase{\textit{et al.}}: Bare Demo of IEEEtran.cls for IEEE Journals}

\maketitle

\begin{abstract}
Recent years have witnessed the remarkable progress of applying deep learning models in video person re-identification (Re-ID). A key factor for video person Re-ID is to effectively construct discriminative and robust video feature representations for many complicated situations. Part-based approaches employ spatial and temporal attention to extract representative local features. While correlations between parts are ignored in the previous methods, to leverage the relations of different parts, we propose an innovative adaptive graph representation learning scheme for video person Re-ID, which enables the contextual interactions between relevant regional features. Specifically, we exploit the pose alignment connection and the feature affinity connection to construct an adaptive structure-aware adjacency graph, which models the intrinsic relations between graph nodes. We perform feature propagation on the adjacency graph to refine regional features iteratively, and the neighbor nodes' information is taken into account for part feature representation. To learn compact and discriminative representations, we further propose a novel temporal resolution-aware regularization, which enforces the consistency among different temporal resolutions for the same identities. We conduct extensive evaluations on four benchmarks, i.e. iLIDS-VID, PRID2011, MARS, and DukeMTMC-VideoReID, experimental results achieve the competitive performance which demonstrates the effectiveness of our proposed method. The code is available at~\url{https://github.com/weleen/AGRL.pytorch}.
\end{abstract}


\IEEEpeerreviewmaketitle

\section{Introduction}\label{sec:introduction}
\begin{figure}[t]
    \begin{center}
        \includegraphics[width=1\linewidth]{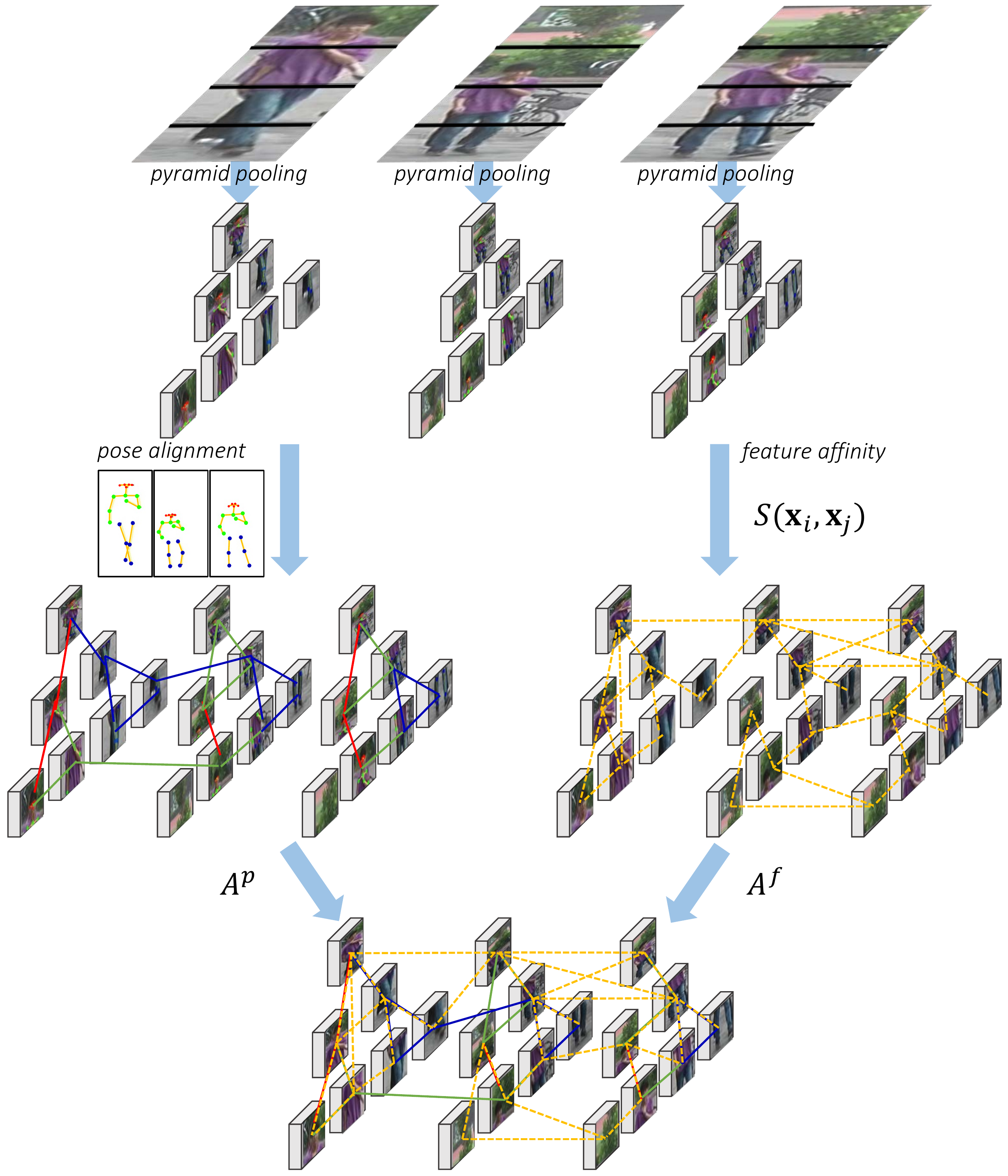}
    \end{center}
    \caption{Overview of graph construction in our proposed method. 1) The feature map from each frame is processed by a pyramid pooling module, the extracted regional features are treated as the graph nodes. 2) The pose alignment adjacency graph $A^p$ (colorful solid line) is constructed by connecting the regions containing the same human part. 3) The feature affinity adjacency graph $A^f$ (yellow dotted line) is constructed by measuring the affinity of regional features. 4) The adaptive structure-aware adjacency graph is built by combining two graphs. \textbf{Best viewed in color, some graph edges are omitted for clarity.}}
\label{fig:graph construction}
\end{figure}

As an important and challenging problem in computer vision, \textit{person re-identification} (Re-ID) aims at precisely retrieving the same identities from the gallery with a person of interest as a query given, and it has a wide range of applications in intelligent surveillance and video analysis~\cite{Zajdel_2005_ICRA}. Typically, person Re-ID is carried out in the domain of individual images without capturing the temporal coherence information. More recently, several video person Re-ID approaches~\cite{McLaughlin_2016_CVPR,Zhou_2017_CVPR,Xu_2017_ICCV,Liu_2017_CVPR,Li_2018_CVPR,Chen_2018_CVPR,Si_2018_CVPR,Dai_2018_TIP,Wu_2019_TNNLS,Zhang_2019_TIP,Chen_2019_TIP,Liu_2019_AAAI,Li_2019_AAAI,Fu_2019_AAAI,Liu_2019_BMVC,Borgia_2019_WACV,Hou_2019_CVPR} emerge to directly perform the person context modeling at the video level, which is more fit for practical use with more visual cues exploited for coping with complicated circumstances.

In the literature, most existing methods for video person Re-ID first extract the feature vectors frame by frame and generate the video-level feature representation by temporal aggregation, then compare them in a particular metric space. Although recent deep learning based methods have made notable progress, re-ID problem remains challenging due to occlusion, viewpoints, illumination, and pose variation in the video. To address these issues, recent studies~\cite{Xu_2017_ICCV,Liu_2017_CVPR,Song_2018_AAAI,Li_2018_CVPR,Wu_2018_TMM} concentrate on aggregating the features from image regions with attention mechanism. However, under the circumstances of complicated situations (e.g. occlusion and pose variations), these approaches are often incapable of effectively utilizing the intrinsic relations between person parts across frames, which play an important role in learning robust video representations. For instance, if the body parts are occluded in the first frame, the appearance cues and contextual information from the other frames are complementary. Hence, how to adaptively perform relation modeling and contextual information propagation among spatial regions is a key issue to solve in video person Re-ID.

Motivated by the aforementioned observations, we propose an adaptive graph learning scheme to model the contextual relations and propagate complementary information simultaneously. As shown in Figure~\ref{fig:graph construction}, we construct two kinds of relations named pose alignment connection and feature affinity connection between the spatiotemporal regions. Specifically, 1) pose alignment connection: regions containing the same part are connected to align the spatial regions across the frames. With the pose alignment connection, we are capable of capturing the relations between human body parts; and 2) feature affinity connection: we define the edges by measuring the visual correlations of extracted regional features. With the feature affinity connection, we can model the visually semantic relationships between regional features precisely.

By combining these two complementary relation connections, we obtain an adaptive structure-aware adjacency graph. Then we capture the contextual interactions on the graph structure via graph neural network (GNN). The effective and discriminative messages aggregated from neighbors are used to refine the original regional features. With feature propagation, the discrimination of the informative regional features is enhanced and the noisy parts are weakened.

Based on the observations in~\cite{Zhang_2018_CVPR}, the visual cues in the video are rich yet possibly redundant, and some keyframes are sufficient to represent the long-range video. As a consequence, we propose a novel regularization, which enforces the consistency among different temporal resolutions, to learn the temporal resolution invariant representation. Specifically, the frame-level features are randomly selected as the subsequences, and input into the attention module, then the output video representations are enforced to be close to each other in the metric space.

Overall, the main contributions of this work are summarized as follows:
\begin{itemize}
    \item We propose an adaptive structure-aware spatiotemporal graph representation based on two types of graph connections for relation modeling: pose alignment connection and feature affinity connection. By combining these two relation connections, the adaptive structure-aware graph representation is capable of well capturing the semantic relations between regions across frames.
    \item We propose a novel regularization to learn the temporal resolution invariant representation, which is compact and captures the discriminative information in the sequence.
\end{itemize}
We conduct extensive experiments on four widely used benchmarks (i.e. iLIDS-VID, PRID2011, MARS, and DukeMTMC-VideoReID), and experimental results demonstrate the effectiveness of our proposed method.

%-------------------------------------------------------------------------
\section{Related Work}\label{sec:related work}
\textbf{Person Re-ID.}
Person Re-ID in still images is widely explored~\cite{Zhao_2017_ICCV,Sun_2018_ECCV,Fuyang_2019_AAAI,Zheng_2019_CVPR,Zhou_2019_ICCV,Yu_2018_ECCV,Zhou_2017_ICCV,Xiao_2017_CVPR,Saquib_2018_CVPR,Qian_2018_ECCV,Su_2017_ICCV}. Currently, the researchers start to focus on video-based person Re-ID~\cite{McLaughlin_2016_CVPR,Yan_2016_ECCV}. Facilitated by deep learning techniques, impressive progress has been observed with video person Re-ID recently. McLaughlin \etal~\cite{McLaughlin_2016_CVPR} and Yan \etal~\cite{Yan_2016_ECCV} employ RNN to model the inter-sequence dependency and aggregate the features extracted from the video frames with average pooling or max pooling. Zhou \etal~\cite{Zhou_2017_CVPR} separately model the spatial and temporal coherence with two RNN networks: the temporal model (TAM) focuses on discriminative frames and spatial model (SRM) integrates the contexture at different locations for better similarity evaluation. Wu \etal~\cite{Wu_2018_TMM} extend GRU with attention mechanism to selectively propagate relevant features and memorize their spatial dependencies through the network. Dai \etal~\cite{Dai_2018_TIP} proposes a $S^2TN$ network to address the pose alignment and combine the bi-directional LSTM with residual learning to perform temporal residual learning.

Recently, the attention networks are widely studied for temporal feature fusion. In~\cite{Xu_2017_ICCV,Liu_2017_CVPR}, the discriminative frames are selected with attention temporal pooling, where each frame is assigned with a quality score and then fused to a final video representation. Similarly, Zhang \etal~\cite{Zhang_2018_CVPR} employ reinforcement learning to train an agent to verify whether the pair of images are the same or different, and the Q value is a good indicator of the difficulty of image pairs. In~\cite{Li_2018_CVPR,Song_2018_AAAI,Liu_2019_AAAI,Fu_2019_AAAI}, the authors extend the temporal attention to spatiotemporal attention to select informative regions and achieve the impressive improvements. Chen \etal~\cite{Chen_2019_TIP} leverage the body joints to attend to the saliency parts of the person in the video to extract the discriminative local features in a siamese network. And in~\cite{Chen_2018_CVPR,Zhang_2019_TIP}, the video representation is generated by considering not only intra-sequence influence but also inter-sequence mutual information. Different from the previous 2D CNN based methods, 3D convolution neural network (3D CNN) is also adopted to address the video person Re-ID~\cite{Wu_2019_TNNLS,Li_2019_AAAI,Liu_2019_BMVC}. Wu \etal~\cite{Wu_2019_TNNLS} adopt 3D CNN and 3D pooling to aggregate the spatial and temporal cues simultaneously. Li \etal~\cite{Li_2019_AAAI} propose a variant of ResNet by inserting multi-scale 3D (M3D) layer and residual attention layer (RAL) into the ResNet. Similarly, Liu \etal~\cite{Liu_2019_BMVC} incorporate non-local modules with ResNet50 as the Non-local Video Attention Network (NVAN) and propose a spatially and temporally efficient variant. Moreover, attributions are utilized to generate the confidence as the weight for sub-features extracted from video frames in~\cite{Zhao_2019_CVPR}. The generative models are adopted to address the occlusion and pose variant in~\cite{Borgia_2019_WACV,Hou_2019_CVPR}.

\begin{figure*}[t]
    \begin{center}
    \includegraphics[width=1\linewidth]{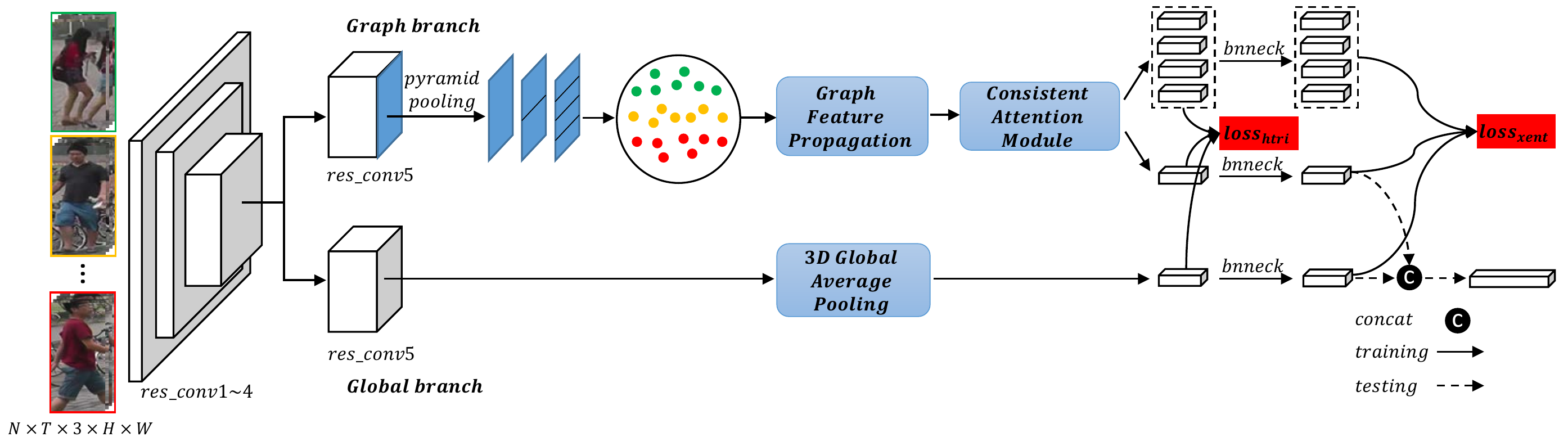}
    \end{center}
    \caption{The overall architecture of our proposed method. 1) $T$ frames are sampled from a long-range video with a restricted random sampling method. 2) In graph branch, for the output of each image, pyramid pooling is used to extract the $N\times d$-dimension feature, where $N$ represents the number of regions, the feature vector for each region has $d$ dimensions. 3) The extracted feature vectors are treated as the graph nodes, we then employ GNN to perform feature propagation on the graph iteratively in the graph feature propagation module. 4) We carry out the attention module to generate the discriminative video representation, the subsequences are randomly selected and forward into the attention module to learn a consistent video representation. 5) Feature vectors from graph branch and global branch are concatenated for testing.}
\label{fig:architecture}
\end{figure*}

\textbf{Graph Models.}
Graph models are utilized in several computer vision tasks, and Graph Neural Networks (GNN) is introduced in~\cite{Scarselli_2009_TNN} to model the relations between graph nodes, and a large number of the variants~\cite{Hamilton_2017_NIPS,Kipf_2017_ICLR,Velickovic_2018_ICLR,Ying_2018_NIPS,Zhang_2018_AAAI,Yan_2018_AAAI} are proposed. In recent, Re-ID methods~\cite{Cheng_2018_PR,Barman_2017_ICCV,Shen_2018_ECCV,Chen_2018_CVPR_Group,Yan_2019_CVPR} combined with graph models are also explored. Cheng \etal~\cite{Cheng_2018_PR} formulate the structured distance relationships into the graph Laplacian form to take advantage of the relationships among training samples. In~\cite{Barman_2017_ICCV}, an algorithm that maps the ranking process to a problem in graph theory is proposed. Shen \etal~\cite{Shen_2018_ECCV} leverage the similarities between different probe-gallery pairs for updating the features extracted from images. Chen \etal~\cite{Chen_2018_CVPR_Group} involves multiple images to model the relationships among the local and global similarities in a unified CRF. Yan \etal~\cite{Yan_2019_CVPR} formulate the person search as a graph matching problem, and solve it by considering the context information in the probe-gallery pairs. To address the unsupervised Re-ID problem, Ye \etal~\cite{Ye_2017_ICCV} involves the graph matching into an iteratively updating procedure for a robust label estimation.

As a closely related problem, there are several graph-based methods~\cite{More_2018_IJAS,Connie_2014_EURASIP,Wu_2019_ACPR}. Connie~\etal~\cite{Connie_2014_EURASIP} propose to learn Grassmannian graph embedding by constraining the geometry of the manifold. In~\cite{More_2018_IJAS}, cross wavelet transform and bipartite graph model are used to extract the dynamic and static features respectively. Our proposed method is similar to~\cite{Wu_2019_ACPR}, which combines the Graph Attention Network (GAT) with the feature extractor to discover the relationship between frames and the variation of a region in the temporal domain, while its region is based on the strong activated point in the feature map and the weight matrix is learned without prior knowledge.

In a nutshell, the graph model based methods in Re-ID usually build up a graph to represent the relationships among training samples, where the graph nodes are images or videos. While in our proposed method, the graph is dynamically learned with prior knowledge to model the intrinsic contextual relationships among the regions in an image sequence, the local, global, and structure information are propagated among the different regional features to learn the discriminative video feature representation.

%-------------------------------------------------------------------------
% our method
\section{The Proposed Method}\label{sec:the proposed method}
\subsection{Overview}\label{sec:overview}
Video person Re-ID aims to retrieve the identities from the gallery with the given queries. The overall architecture of our proposed method is illustrated in Figure~\ref{fig:architecture}. Given a long-range video for the specific identity, $T$ frames are randomly sampled with a restricted sampling method~\cite{Zolfaghari_2018_ECCV,Li_2018_CVPR}, and then grouped as an image sequence $\{I_t\}_{t=1,...,T}$. We first feed them into the ResNet50-based~\cite{He_2016_CVPR} feature extractor module, in which the stride of the first residual block in $conv5$ is set to 1. In the global branch, 3D global average pooling is used for the feature maps and produces a video representation $\mathbf{x}_{gap}\in \mathbb{R}^d$. In the graph branch, we obtain regional features $\mathbf{X}=\{\mathbf{x}_i\}_{i=1}^{T\cdot N}$ with pyramid pooling~\cite{Fuyang_2019_AAAI}, where the feature maps are vertically partitioned into 1, 2, and 4 regions\footnote{In this paper, regions and nodes are interchangeably for the same meaning.} in our experiments, and $N=7$ is the number of regions for an individual frame. Then, we utilize the pose information and feature affinity to construct an adaptive structure-aware adjacency graph, which captures the intrinsic relations between these regional features. In the graph feature propagation module, the regional features are updated iteratively by aggregating the contextual information from neighbors on the graph. Next, we utilize the attention module to yield the video representations. The network is supervised by identification loss and triplet ranking loss together. We will discuss these modules in the following sections.

\subsection{Adaptive Graph Feature Propagation}\label{sec:feature propagation}
As discussed in Section~\ref{sec:introduction}, the relations between human parts are beneficial for mitigating the impact of complex situations such as occlusion and clutter background. So, how to describe the relationships between different human parts and propagate contextual messages is critical for learning the discriminative video representations. The graph is commonly used to model this kind of relations, and we adopt GNN to leverage the information from the neighborhood.\\

\noindent\textbf{Adaptive Structure-Aware Adjacency Graph.}
To depict the relations of human parts, we employ the pose information and feature affinity to construct an adaptive structure-aware adjacency graph $G=\{V, A\}$. $V=\{v_i\}_{i=1}^{T \cdot N}$ is the vertex set containing $T\cdot N$ nodes, where each node $v_i$ corresponds to a spatial region in the frame. To define the edge $A\in\mathbb{R}^{(T\cdot N) \times (T\cdot N)}$ on the graph, we introduce two types of relations: pose alignment connection and feature affinity connection.

\begin{figure}[t]
    \centering
    \includegraphics[width=1\linewidth]{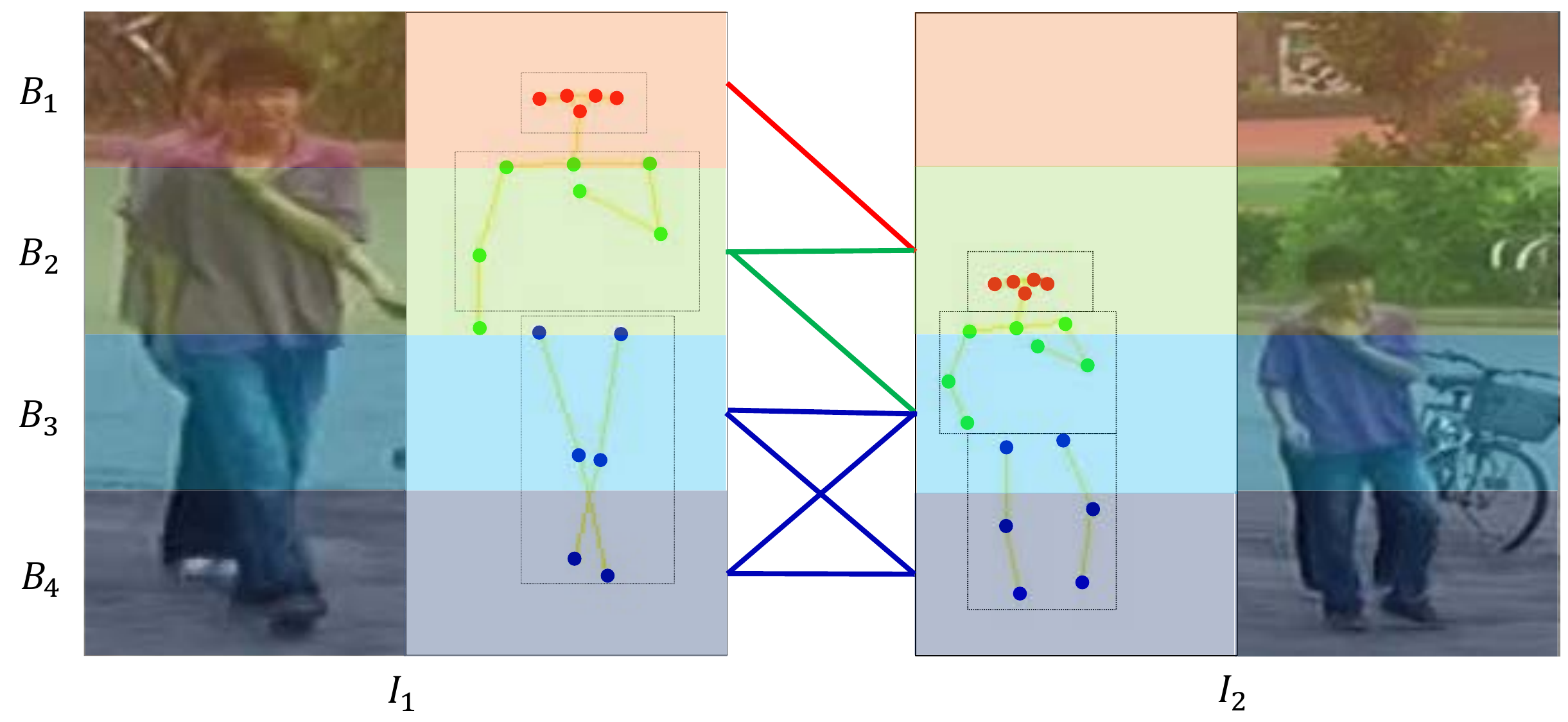}
   \caption{Construction of pose alignment adjacency graph for spatial regions. The estimated keypoints are separated into three parts: head part (consist of nose, neck, eyes, and ears), trunk part (consist of shoulders, bows, and wrists), and leg part (consist of hips, knees, and ankles). The spatial regions (nodes) containing the same part are connected in the pose alignment adjacency graph.}
   %\vspace{-1em}
\label{fig:pose}
\end{figure}

The pose alignment connection is defined by leveraging the human body joints: two regions (nodes) are connected if they contain the same human parts. Formally, we define a set $S_i$ for each region $v_i$ where $S_i \subseteq \{head, trunk, leg\}$. The pose alignment adjacency graph $A^p$ for the two nodes $v_i$ and $v_j$ is then calculated as follows:
\begin{equation}
A^p_{ij} = \left\{
    \begin{aligned}
         & 1 \:\:\:\:\:\:\:i \neq j\:\:\: \text{and} \:\: |S_i\cap S_j| \neq 0,\\
         & 0 \:\:\:\:\:\:\:\text{otherwise},
    \end{aligned}
    \right.
\end{equation}
where $|\cdot|$ means the cardinality of a set. We obtain $S_i$ with the following procedure, first, we locate the joints of the human body by making use of a human pose estimation algorithm, this is illustrated in Figure~\ref{fig:pose}. Then we separate these estimated keypoints into three parts: head part (consist of nose, neck, eyes, and ears), trunk part (consist of shoulders, bows, and wrists), and leg part (consist of hips, knees, and ankles). For the $i$-th spatial region, $S_i$ is constitutive of the parts located in this spatial region. We present an example in Figure~\ref{fig:pose}, the feature maps are vertically partitioned into 4 regions $B_1, B_2, B_3, B_4$. In image $I_1$ and $I_2$, the head part is in $B_1$ and $B_2$ respectively, so the pose alignment connection between these two nodes is set to 1. Then, we can create the pose alignment adjacency graph $A^p$.

Pose alignment adjacency connection reflects only the coarse relations between different spatial regions and the recent method~\cite{Simonovsky_2017_CVPR} shows the dynamic graph could learn better graph representations compared with the fixed graph structure. To describe the fine relations between the regions, we propose to learn an adaptive feature affinity adjacency graph $A^{f}$, which aims to capture the affinity between the regions. For two nodes $v_i$ and $v_j$, the node features are $\mathbf{x}_i$ and $\mathbf{x}_j$ respectively, then the entry of adjacency graph $A^f$ is formulated as follows:
\begin{equation}
    \begin{aligned}
    A^f_{ij} = &S(\mathbf{x}_i, \mathbf{x}_j) \\
             = &\frac{2}{e^{\|\mathbf{x}_i - \mathbf{x}_j\|_2} + 1}.
    \end{aligned}
\label{eq:feature affinity}
\end{equation}

We calculate the edge weight matrix $A$ by combining the pose alignment adjacency matrix and feature affinity matrix:
\begin{equation}
    A_{ij} = \frac{1}{1 + \gamma} (\frac{A^{p}_{ij}}{\sum_{j}{A^{p}_{ij}}} + \gamma \frac{A^f_{ij}}{\sum_{j}{A^f_{ij}}}),
\end{equation}
where $\gamma$ is the weight parameter to balance the pose alignment adjacency matrix and the feature affinity matrix, and $\gamma$ is set to 1 in our all experiments.\\

\noindent\textbf{Graph Feature Propagation Module.}
\begin{figure}[t]
     \centering
     \includegraphics[width=0.8\linewidth]{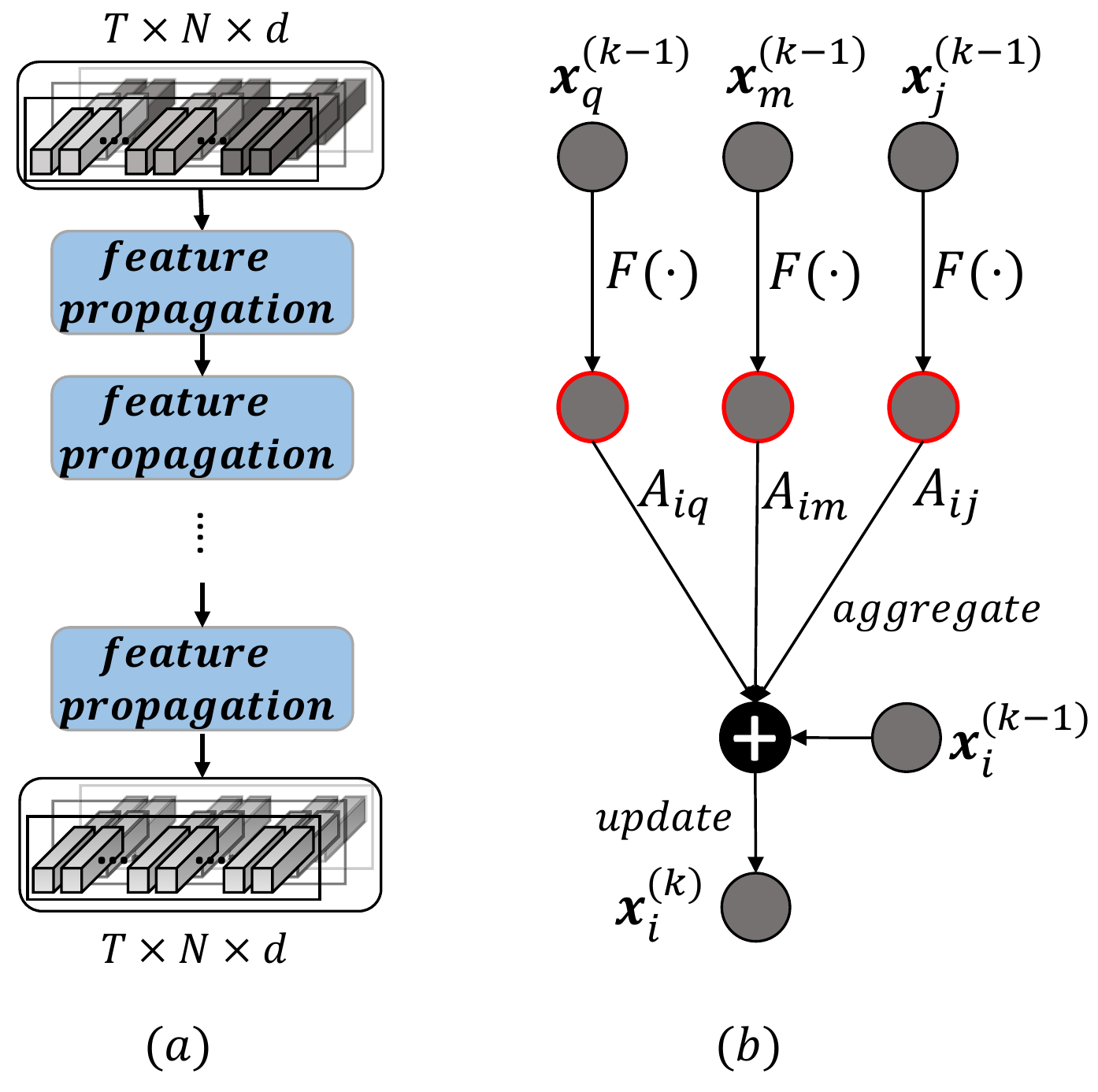}
    \caption{(a) Graph feature propagation module. Given the adaptive pose alignment adjacency graph $A$, the original spatial regional features are updated iteratively through the feature propagation layers. (b) Feature propagation layer. The features from neighbors are processed by an fc layer $F(\cdot)$, and then aggregated by the weights from the adjacency graph. The operation is defined in Equation~\ref{eq:aggregation}.}
\label{fig:graph propagation}
\end{figure}
After obtaining the graph, we perform contextual message propagation to update original spatial regional features iteratively.

As shown in Figure~\ref{fig:graph propagation}, we employ Graph Neural Network (GNN)~\cite{Scarselli_2009_TNN} to aggregate the information from neighbors for each node. In the graph feature propagation module, we stack $L$ graph feature propagation layers, in the $l$-th layer, the aggregation and updating operations are defined as follows:
\begin{equation}
\begin{aligned}
&\mathbf{x}_i^{(l)} = (1 - \alpha)\mathbf{x}_i^{(l-1)} + \alpha \sum_{j = 1}^{T\cdot N}A_{ij}^{(l)} F^{(l)}(\mathbf{x}_j^{(l-1)})
\end{aligned}\label{eq:aggregation}
\end{equation}
where $i\in\{1,2,\dots,T\cdot N\}$, $l\in\{1,2,\dots,L\}$, and $\mathbf{x}_i^{(l)}$ stands for the refined regional feature output from $l$-th feature propagation layer and $\mathbf{x}_i^{(0)}=\mathbf{x}_i$ is the original node feature, $F^{(l)}(\cdot)$ is the combination of an FC-layer and batch normalization layer to encode the contextual messages from neighbors, $A^{(l)}$ refers to the adaptive structure-aware adjacency graph, and $\alpha$ is used to balance the aggregated feature and original feature, which is set as $0.1$ in our experiments. The output from graph feature propagation module is denoted as $\mathbf{\hat{X}} = [\mathbf{\hat{x}}_1, \mathbf{\hat{x}}_2, \dots, \mathbf{\hat{x}}_{T\cdot N}]$, where $\mathbf{\hat{x}}_i \in \mathbb{R}^d$ is the updated regional feature vector.

%%-------------------------------------------------------------------------
%% temporal attention module
\noindent\textbf{Temporal Attention Module}\label{sec:temporal attention module}
Given the updated regional features $\mathbf{\hat{X}}$, we perform a simple yet effective spatio-temporal attention~\cite{Fu_2019_AAAI} to obtain the video representation, which is calculated as:
\begin{equation}
    \begin{aligned}
    \mathbf{x}_{graph} = \sum_{i=1}^{T\cdot N} \frac{\|\mathbf{\hat{x}}_i\|_1}{\sum_j{\|\mathbf{\hat{x}}_j\|_1}}\mathbf{\hat{x}}_i.
    \end{aligned}
\end{equation}
As discussed in Section~\ref{sec:introduction}, in order to model the consistency of subsequences, we randomly select $T-i$ frames from the image sequence, where $i={1, 2, \dots, T_s}$. And then feed the feature vectors of these subsequence frames into the temporal attention module to obtain video representations $\mathbf{x}_{graph,1}, \dots, \mathbf{x}_{graph,T_s}$. To keep the consistency of different subsequences, we enforce these video representations to be close to each other in the metric space.

%%-------------------------------------------------------------------------
%% optimization
\subsection{Loss Functions}\label{sec:optimization}
We employ two kinds of losses to jointly supervise the training of parameters: cross entropy loss and soft hard triplet loss~\cite{Hermans_2017_ArXiv}, the losses are formulated as follows:
\begin{equation}
    L_{xent}(\mathbf{x}) = -\frac{1}{P\cdot K}\sum_{i=1}^{P}\sum_{a=1}^{K} log[\frac{e^{{\mathbf{W}_{y_{i,a}}}^{\top}\mathbf{x}_{i,a}}}{\sum_{c=1}^{P \cdot K} e^{{\mathbf{W}_{c}}^{\top}\mathbf{x}_{i,a}}}],
\end{equation}
\begin{equation}
    \begin{aligned}
        L_{htri}(\mathbf{x}) = \sum_{i=1}^P\sum_{a=1}^{K}ln(1+&exp(
        \overbrace{\max_{p=1,\cdots, K} D(\mathbf{x}_{i,a}, \mathbf{x}_{i,p})}^{hardest \:\: positive} \\
       &-\underbrace{\min_{\begin{subarray}{c} n=1,...,K \\
                                               j=1,...,P \\
                                               j \neq i
                                                \end{subarray}} D(\mathbf{x}_{i,a}, \mathbf{x}_{j,n})}_{hardest \:\: negative})),
    \end{aligned}
\end{equation}
where $P$ and $K$ are respectively the number of identities and sampled images of each identity. So there are $P\cdot K$ images in a mini-batch, $\mathbf{x}_{i, a}$, $\mathbf{x}_{i, p}$ and $\mathbf{x}_{j, n}$ are the features extracted from the anchor, positive and negative samples respectively, $D(\cdot)$ is the L2-norm distance for two feature vectors.

For the output from global branch, we have two losses $l_{xent}^{global}$ and $l_{htri}^{global}$:
\begin{equation}
    l_{xent}^{global} = L_{xent}([BN(\mathbf{x}^{(1)}_{gap}), \dots, BN(\mathbf{x}^{(P\cdot K)}_{gap}])),
\end{equation}
\begin{equation}
l_{htri}^{global} = L_{htri}([\mathbf{x}^{(1)}_{gap}, \dots, \mathbf{x}^{(P\cdot K)}_{gap}]),
\end{equation}
for the output from graph branch, we have two losses $l_{xent}^{graph}$ and $l_{htri}^{graph}$ similarily:
\begin{equation}
    \begin{aligned}
    l_{xent}^{graph} = L_{xent}([BN(\mathbf{x}^{(1)}_{graph}), \dots, BN(\mathbf{x}^{(1)}_{graph, T_s}),\\
     \dots, BN(\mathbf{x}^{(P\cdot K)}_{graph}), \dots, BN(\mathbf{x}^{(P\cdot K)}_{graph, T_s})]),
    \end{aligned}
\end{equation}
\begin{equation}
    \begin{aligned}
    l_{htri}^{graph} = L_{htri}([\mathbf{x}^{(1)}_{graph}, \mathbf{x}^{(1)}_{graph,1}, \dots, \mathbf{x}^{(1)}_{graph, T_s}, \\\dots, \mathbf{x}^{(P\cdot K)}_{graph}, \mathbf{x}^{(P\cdot K)}_{graph,1}, \dots, \mathbf{x}^{(P\cdot K)}_{graph, T_s}]),
    \end{aligned}
\end{equation}
where $BN(\cdot)$ is the BNNeck introduced in~\cite{Luo_2019_CVPRW}, $[\cdot]$ means concatenation. The total loss is the summation of the four losses:
\begin{equation}
l_{total} = l_{xent}^{global} + l_{htri}^{global} + l_{xent}^{graph} + l_{htri}^{graph}
\end{equation}

%-------------------------------------------------------------------------
% experiments
\section{Experiments}\label{sec:experiments}
\subsection{Datasets}
\textbf{PRID2011}~\cite{Wang_2014_ECCV} dataset consists of person videos from two camera views, containing 385 and 749 identities, respectively. Only the first 200 people appear in both cameras. The length of the image sequence varies from 5 to 675 frames, but we use only the sequences whose frame number is larger than 21. \\
\indent \textbf{iLIDS-VID}~\cite{Hirzer_2011_SCIA} dataset consists of 600 image sequences of 300 persons. For each person, there are two videos with the sequence length ranging from 23 to 192 frames with an average duration of 73 frames. \\
\indent \textbf{MARS} dataset~\cite{Zheng_2016_ECCV} is the largest video-based person re-identification benchmark with 1,261 identities and around 20,000 video sequences generated by DPM detector and GMMCP tracker. The dataset is captured by six cameras, each identity is captured by at least 2 cameras and has 13.2 sequences on average. There are 3,248 distracter sequences in the dataset, it increases the difficulty of Re-ID.\\
\indent \textbf{DukeMTMC-VideoReID} dataset is another large scale benchmark dataset for video-based person Re-ID, which is derived from DukeMTMC dataset~\cite{Ristani_2016_ECCVW} and re-organized by Wu \etal~\cite{Wu_2018_CVPR}. DukeMTMC-VideoReID dataset contains a total of 4,832 tracklets and 1,812 identities, it is separated into 702, 702, and 408 identities for training, testing, and distraction. In total, it has 369,656 frames of 2,196 tracklets for training and 445,764 frames of 2,636 tracklets for testing and distraction. Each tracklet has 168 frames on average, and the bounding boxes are annotated manually.

\subsection{Evaluation Metrics}
For evaluation, we employ the standard metrics used in person Re-ID literature: cumulative matching characteristic (CMC) curve and mean average precision (mAP). CMC curve judges the ranking capabilities of the Re-ID model, mAP reflects the true ranking results while multiple ground-truth sequences exist. For PRID2011 and iLIDS-VID datasets, we follow the evaluation protocol used in~\cite{Wang_2014_ECCV}. Each dataset is divided into two parts for training and testing, the final accuracy is the average of ``10-fold cross validation'', only CMC accuracy is reported in PRID2011 and iILIDS-VID because of the equivalence of CMC and mAP on these two datasets. For MARS and DukeMTMC-VideoReID dataset, both CMC and mAP are reported.

\subsection{Implementation Details}~\label{sec:implement}
\textbf{Settings.}
Our experiments are implemented with Pytorch and four TITAN X GPUs. ResNet50~\cite{He_2016_CVPR} is first pre-trained on ImageNet, and the input images are all resized to $256\times128$. In the training stage, we employ a restricted random sampling strategy~\cite{Li_2018_CVPR} to randomly sample $T=8$ frames from every video and group them into a tracklet. We update the parameters by employing ADAM~\cite{Kingma_2015_ICLR} with a learning rate of $1\times 10^{-4}$ and weight decay of $5\times 10^{-4}$. We train the network for 300 epochs, the learning rate decays to $\frac{1}{10}$ every 100 epochs. For batch hard triplet loss, we set $P=4$ and $K=4$ in our experiments. In the temporal attention module, $T_s$ is set as 3.
In the testing stage, cosine distance between the representations is calculated for ranking, a video containing $T_v$ frames is split into $T$ chunks firstly, then we make use of two kinds of strategies: 1) the first image is collected as an image sequence to represent this video; 2) In each chunk, $i$-th frames are grouped as an image sequence, we can obtain $\lceil \frac{T_v}{T} \rceil$ image sequences, and the video representations are averaged as a single video representation. In our experiments, the first strategy is fast and the second strategy is more accurate.

\textbf{Pose Estimation.}
In our implementation, the pose information is obtained by AlphaPose~\cite{Su_2017_ICCV} before the training. For the cases that multiple persons exist in the frame, we get the pose following the criterion: 1) The bounding box for the correct identity is bigger than the other persons; 2) The correct identity is in the center of the image, here we use the detected center point to represent the person; 3) The average confidence score of the keypoints for the correct identity is higher than the others.
Then we define the body parts heuristically based on the pose information, and we construct the pose alignment graph which builds up the relations between the vertically partitioned regional feature vectors. Combine the pose alignment graph with the feature affinity graph, we refine the feature vectors in the feature propagation module and utilize the refined feature vectors for Re-ID task.

\begin{figure}[t]
    \centering
    \includegraphics[width=1\linewidth]{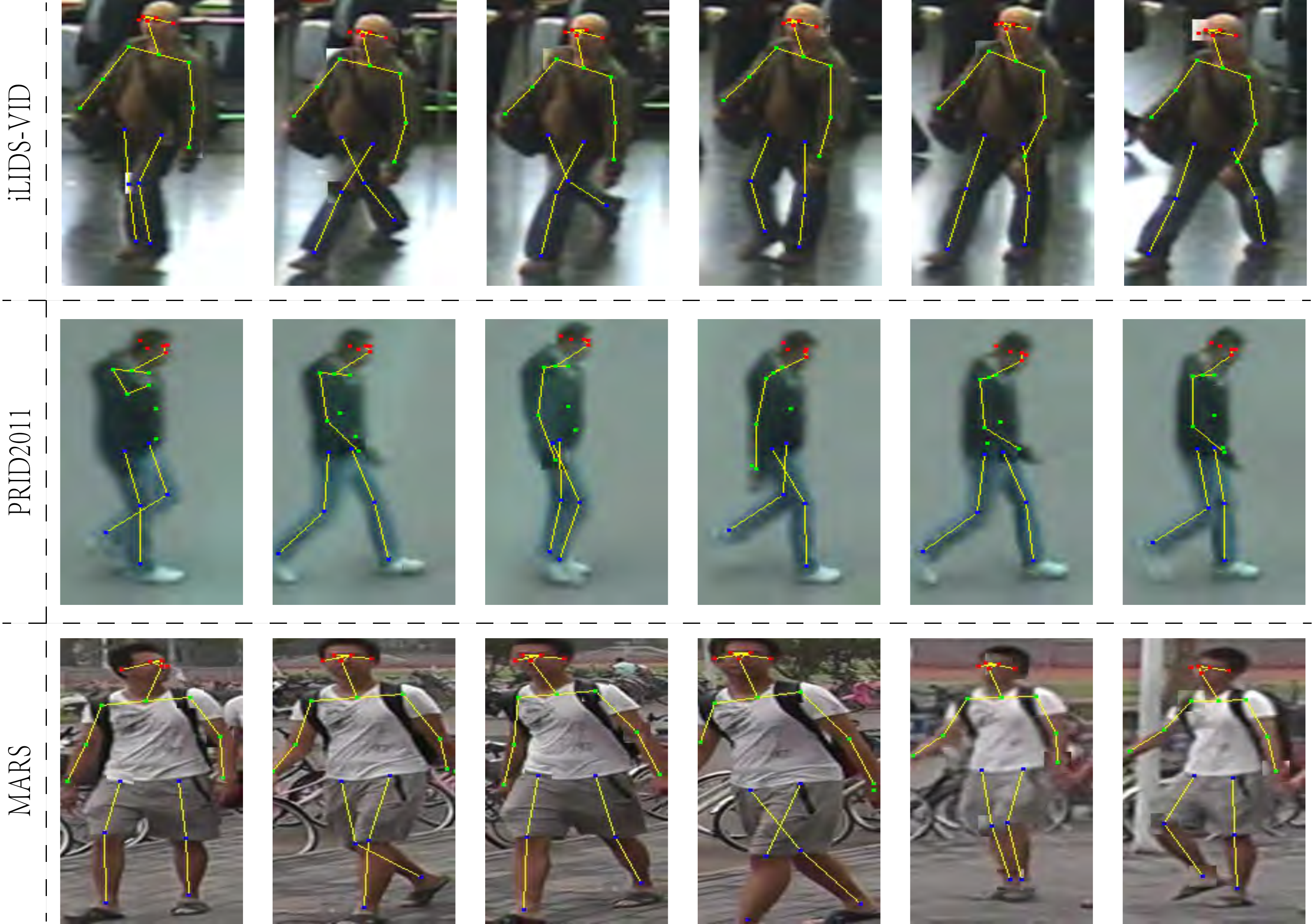}
    \caption{Keypoint detection examples. The human body joints are marked by colorful dots, and linked by yellow lines. \textbf{Best viewed in color and zoom in.}}
    \label{fig:keypoints}
\end{figure}

\subsection{Comparison with State-of-the-art Methods}
To validate the effectiveness of our proposed method, we compare our proposed method with several state-of-the-art methods on PRID2011, iLIDS-VID, MARS, and DukeMTMC-VideoReID, include RCN~\cite{McLaughlin_2016_CVPR}, IDE+XQDA~\cite{Zheng_2016_ECCV}, RFA-Net~\cite{Yan_2016_ECCV}, SeeForest~\cite{Zhou_2017_CVPR}, QAN~\cite{Liu_2017_CVPR}, AMOC+EF~\cite{Liu_2017_TCSVT}, ASTPN~\cite{Xu_2017_ICCV}, Snippet~\cite{Chen_2018_CVPR}, STAN~\cite{Li_2018_CVPR}, EUG~\cite{Wu_2018_CVPR}, SDM~\cite{Zhang_2018_CVPR}, RQEN~\cite{Song_2018_AAAI}, PersonVLAD~\cite{Wu_2019_TNNLS}, M3D~\cite{Li_2019_AAAI}, STMP~\cite{Liu_2019_AAAI}, STA~\cite{Fu_2019_AAAI}, TRL+XQDA~\cite{Dai_2018_TIP}, SCAN~\cite{Zhang_2019_TIP}, STAL~\cite{Chen_2019_TIP}, and STE-NVAN~\cite{Liu_2019_BMVC}.

\begin{table}[ht]
    \centering
        \begin{tabular}{ l | c | c | c | c }
            \hline
            \multirow{2}{*}{Method} & \multicolumn{4}{c}{MARS} \\\cline{2-5}
            & R1 & R5 & R20 &mAP\\ \hline
            IDE+XQDA~\cite{Zheng_2016_ECCV} & 65.3 & 82 & 89 & 47.6\\
            SeeForest~\cite{Zhou_2017_CVPR} & 70.6 & 90 & 97.6 & 50.7\\
            ASTPN~\cite{Xu_2017_ICCV} & 44 & 70 & 81 & - \\
            RQEN~\cite{Song_2018_AAAI} & 73.74 & 84.9 & 91.62 & 71.14 \\
            Snippet~\cite{Chen_2018_CVPR} & 81.2 & 92.1 & - & 69.4\\
            STAN~\cite{Li_2018_CVPR} & 82.3 & - & - & 65.8\\
            SDM~\cite{Zhang_2018_CVPR} & 71.2 & 85.7 & 94.3 & - \\
            EUG\cite{Wu_2018_CVPR} & 62.67 & 74.94 & 82.57 & 42.45 \\
            PersonVLAD~\cite{Wu_2019_TNNLS} & 80.8 & 94.5 & \textbf{99} & 63.4 \\
            % PersonVLAD+XQDA~\cite{Wu_2019_TNNLS} & 82.8 & 94.9 & \textbf{99} & 64.7 \\
            DSAN+KISSME~\cite{Wu_2018_TMM} & 73.5 & 85 & 97.5 &  \\
            TRL+XQDA~\cite{Dai_2018_TIP} & 80.5 & 91.8 & 96 & 69.1\\
            M3D~\cite{Li_2019_AAAI} & 84.39 & 93.84 & 97.74 & 74.06 \\
            STA~\cite{Fu_2019_AAAI} & 86.3 & 95.7 & 98.1 & 80.8 \\
            STMP~\cite{Liu_2019_AAAI} & 84.4 & 93.2 & 96.3 & 72.7 \\
            SCAN~\cite{Zhang_2019_TIP} & 86.6 & 94.8 & 97.1 & 76.7 \\
            STAL~\cite{Chen_2019_TIP} & 82.2 & 92.8 & 98 & 73.5 \\
            % VRSTC~\cite{Hou_2019_CVPR} & 88.5 & 96.5 & 97.4 & \textbf{82.3} \\
            STE-NVAN~\cite{Liu_2019_BMVC} & 88.9 & - & - & 81.2 \\\hline
            AMOC+EF\dag~\cite{Liu_2017_TCSVT} & 68.3 & 81.4 & 90.6 & 52.9 \\
            Snippet+OF\dag~\cite{Chen_2018_CVPR} & 86.3 & 94.7 & - & 76.1\\
            SCAN+OF\dag~\cite{Zhang_2019_TIP} & 87.2 & 95.2 & 98.1 & 77.2 \\\hline
            Ours & \textbf{89.8} & 96.1 & 97.6 & 81.1 \\
            \hspace{1em}+Test Strategy 2 & 89.5 & \textbf{96.6} & 97.8 & \textbf{81.9} \\\hline
        \end{tabular}
        \vspace{0.5em}
    \caption{Comparison with state-of-the-art methods on MARS dataset, Rank-1, -5, -20 accuracies(\%) and mAP are reported. $\dag$ refers to optical flow, and \textit{Test Strategy 2} is the second strategy introduced in Section~\ref{sec:implement}.}
    \label{tab:mars}
\end{table}

\begin{table}[ht]
    \centering
        \begin{tabular}{ l | c | c | c | c }
            \hline
            \multirow{2}{*}{Method} & \multicolumn{4}{c}{DukeMTMC-VideoReID} \\\cline{2-5}
            & R1 & R5 & R20 & mAP\\ \hline
            STA \cite{Fu_2019_AAAI} & 96.2 & \textbf{99.3} & - & 94.9 \\
            STE-NVAN \cite{Liu_2019_BMVC} & 95.2 & - & - & 93.5 \\
            EUG\cite{Wu_2018_CVPR} & 83.6 & 94.6 & 97.6 & 78.3 \\
            VRSTC \cite{Hou_2019_CVPR}& 95 & 99.1 & - & 93.5 \\ \hline
            Ours & 96.7 & 99.2 & 99.7 & 94.2 \\
            \hspace{1em}+Test Strategy 2 & \textbf{97.0} & \textbf{99.3} & \textbf{99.9} & \textbf{95.4} \\ \hline
        \end{tabular}
        \vspace{0.5em}
    \caption{Comparison with state-of-the-art methods on DukeMTMC-VideoReID dataset, Rank-1, -5, -20 accuracies(\%) and mAP are reported. $\dag$ refers to optical flow, and \textit{Test Strategy 2} is the second strategy introduced in Section~\ref{sec:implement}.}
    \label{tab:dukev}
\end{table}
\textbf{Results on MARS and DukeMTMC-VideoReID}
From Table~\ref{tab:mars} and Table~\ref{tab:dukev}, it is not difficult to find that our proposed method outperforms the existing approaches. On MARS dataset, our proposed method surpasses the previous best approach STE-NVAN~\cite{Liu_2019_BMVC} by 0.6\% and 0.7\% in terms of Rank-1 and mAP. And on DukeMTMC-VideoReID, our method achieves the best performance with 97.0\% and 95.4\% at Rank-1 and mAP accuracy respectively. The results outperform the state-of-the-art STA~\cite{Fu_2019_AAAI} by a large margin. These experimental results confirm the effectiveness and superiority of our proposed method.

\begin{table}[t]
    \centering
        \begin{tabular}{ l | c | c | c | c | c | c  }
            \hline
            \multirow{2}{*}{Method} & \multicolumn{3}{c|}{iLIDS-VID} &\multicolumn{3}{c}{PRID2011} \\\cline{2-7}
            & R1 & R5 & R20       & R1 & R5 & R20 \\ \hline
            IDE+XQDA~\cite{Zheng_2016_ECCV} & 53 & 81.4 & 95.1 & 77.3 & 93.5 & 99.3 \\
            RFA-Net~\cite{Yan_2016_ECCV} & 58.2 & 85.8 & 97.9 & 49.3 & 76.8 & 90 \\
            RCN~\cite{McLaughlin_2016_CVPR} & 58 & 84 & 96 & 70 & 90 & 97 \\
            SeeForest~\cite{Zhou_2017_CVPR} & 55.2 & 86.5 & 97 & 79.4 & 94.4 & 99.3 \\
            QAN~\cite{Liu_2017_CVPR} & 68 & 86.8 & 97.4 & 90.3 & 98.2 & \textbf{100} \\
            ASTPN~\cite{Xu_2017_ICCV} & 62 & 86 & 98 & 77 & 95 & 99 \\
            % STFV3D+KISSME~\cite{Zhang_2017_TIP} & 44.3 & 71.7 & 91.7 & 64.1 & 87.3 & 92 \\
            RQEN~\cite{Song_2018_AAAI} & 76.1 & 92.9 & 99.3 & 92.4 & 98.8 & \textbf{100} \\
            Snippet~\cite{Chen_2018_CVPR} & 79.8 & 91.8 & - & 88.6 & 99.1 & - \\
            STAN~\cite{Li_2018_CVPR} & 80.2 & - & - & 93.2 & - & - \\
            SDM~\cite{Zhang_2018_CVPR} & 60.2 & 84.7 & 95.2 & 85.2 & 97.1 & 99.6 \\
            DSAN+KISSME~\cite{Wu_2018_TMM} & 61.9 & 86.8 & 98.6 & 77 & 96.4 & 99.4 \\
            TRL+XQDA~\cite{Dai_2018_TIP} & 57.7 & 81.7 & 94.1 & 87.8 & 97.4 & 99.3 \\
            M3D~\cite{Li_2019_AAAI} & 74 & 94.33 & - & 94.4 & \textbf{100} & - \\
            STMP~\cite{Liu_2019_AAAI} & 84.3 & 96.8 & \textbf{99.5} & 92.7 & 98.9 & 99.8 \\
            PersonVLAD~\cite{Wu_2019_TNNLS} & 69.4 & 87.6 & 99.2 & 87.6 & 96.1 & 99.8 \\
            %PersonVLAD+XQDA~\cite{Wu_2019_TNNLS} & 70.7 & 88.2 & 99.2 & 88 & 96.2 & 99.7 \\
            SCAN~\cite{Zhang_2019_TIP} & 81.3 & 93.3 & 98 & 92 & 98 & \textbf{100} \\
            STAL~\cite{Chen_2019_TIP} & 82.8 & 95.3 & 98.8 & 92.7 & 98.8 & \textbf{100} \\
            VRSTC~\cite{Hou_2019_CVPR} & 83.4 & 95.5 & 99.5 & - & - & - \\ \hline
            AMOC+EF\dag~\cite{Liu_2017_TCSVT} & 68.7 & 94.3 & 99.3 & 83.7 & 98.3 & 100 \\
            Snippet+OF \dag~\cite{Chen_2018_CVPR} & 85.4 & 96.7 & - & 93 & 99.3 & - \\
            SCAN+OF \dag~\cite{Zhang_2019_TIP} & 88 & 96.7 & 100 & 95.3 & 99 & 100 \\\hline
            Ours & 83.7 & 95.4 & 99.5 & 93.1 & 98.7 & 99.8 \\
            \hspace{1em}+Test Strategy 2& \textbf{84.5} & 96.7 & \textbf{99.5} & \textbf{94.6} & 99.1 & \textbf{100} \\\hline
        \end{tabular}
        \vspace{0.5em}
    \caption{Comparison with state-of-the-art methods on PRID2011 and iLIDS-VID datasets, Rank-1, -5, -20 accuracies(\%) are reported. $\dag$ refers to optical flow, and \textit{Test Strategy 2} is the second strategy introduced in Section~\ref{sec:implement}. The approaches utilizing optical flow are not directly compared. The experimental results indicate that our proposed method achieves the state-of-the-art performance.}
    %\vspace{-2em}
    \label{tab:ilids and prid}
\end{table}

\textbf{Results on iLIDS-VID and PRID2011}
As shown in Table~\ref{tab:ilids and prid}, results demonstrate the advantages of the proposed method over existing state-of-the-art approaches on iLIDS-VID and PRID2011. Specifically, our proposed method achieves the Rank-1 accuracy of 84.5\% and 94.6\% on these two datasets, and surpass all the previous approaches without incorporating optical flow. On these two small-scale datasets, by comparing Snnipet and Snippet+OF, we can observe that the motion information provides more reliable features than appearance cues. While even compared with those methods utilizing optical flow, the results of our proposed method are also competitive.

%We think the reason why our results are inferior than the optical flow based approaches may be: 1) The training samples in these two datasets are limited; 2) The quality of video in the dataset is lower than MARS and DukeMTMC-VideoReID, thus the motion information may provide more reliable features than appearance cues.

\subsection{Ablation Study}
To analyze the effectiveness of components in our proposed method, we conduct several experiments on MARS dataset. The experimental results are summarized in Table~\ref{tab:ablation} and Table~\ref{tab:k}.

\begin{table}[ht]
    \centering
        \begin{tabular}{ l | c | c | c | c | c }
            \hline
            $K$ & R1 & R5 & R10 & R20 & mAP \\ \hline
            1  & 89.3 & 96.0 &97.1  &\textbf{97.6} &80.8  \\
            2  & \textbf{89.8} & \textbf{96.1} &97.0 &\textbf{97.6} & \textbf{81.1}  \\
            3  & 89.5 & 96.1 &\textbf{97.1} &97.0 & \textbf{81.1} \\ \hline
        \end{tabular}
        \vspace{0.5em}
        \caption{Analysis on feature propagation module. $K$ is the number of feature propagation layers. We use the model trained with $K=2$ in our experiments.}
        %\vspace{-2em}
        \label{tab:k}
\end{table}

\begin{table}[ht]
    \centering
        \begin{tabular}{ l | c | c | c | c | c}
            \hline
                Model   & R1 & R5 & R10 &R20 & mAP \\ \hline
                Baseline  & 87.8  &95.3   & 96.3   &97.3 & 78.0 \\
                \hspace{1em}+Attention & 88.0  &95.3  &96.9  &\textbf{97.9} & 79.4 \\
                \hspace{1em}+$\text{A}^p$ & 88.7  & 95.9 & 96.9 &97.8 & 79.8 \\
                \hspace{1em}+$\text{A}^f$ & 88.5  & 95.9 & 97.1 &98.0 & 79.8 \\
                \hspace{1em}+$\text{A}^p$+$\text{A}^f$  & 89.3 &96.1 & 96.9 &97.8 &80.4\\
                \hspace{1em}+$\text{A}^p$+$\text{A}^f$+consistent  & \textbf{89.8}  & \textbf{96.1}  &\textbf{97.0} &97.6 & \textbf{81.1} \\ \hline
        \end{tabular}
        \vspace{0.5em}
        \caption{Ablation study on MARS dataset, we present Rank-1, -5, -10, -20 accuracy(\%) and mAP(\%). $\text{A}^p$, $\text{A}^f$, $\text{A}^p$+$\text{A}^f$, and consistent represent pose alignment adjacency graph, feature affinity graph, combined adjacency graph, and consistent loss respectively. Baseline consists of feature extractor and temporal average pooling, the number of feature propagation layer is set to 2.}
        %\vspace{-2em}
        \label{tab:ablation}
\end{table}

\textbf{Analysis on feature propagation module.}
We carry out experiments to investigate the effect of varying the number of feature propagation layers. We evaluate the results of stacking 1, 2, and 3 propagation layers based on the model combined with adaptive pose alignment adjacency graph and consistent loss, i.e. $+A^p+A^f+consistent$ in Table~\ref{tab:ablation}. As shown in Table~\ref{tab:k}, we find out that the performance is consistent in these settings, where Rank-1 accuracies are all above 89.0\% and mAP accuracies are all above 80.5\%. In addition, the performance with $K=2$ surpasses the other settings, so we stack 2 propagation layers in our experiments.

\textbf{Analysis on components.} In Table~\ref{tab:ablation}, \textit{Baseline} contains only the ResNet backbone and 3D global average pooling, and is supervised by $l_{xent}^{gap}$ and $l_{htri}^{gap}$, the Rank-1 and mAP accuracy of baseline approach is 87.8\% and 78.0\% respectively. \textit{+Attention} refers to adopt temporal attention module in the graph branch, the corresponding performance is 88.0\% in Rank-1 and 79.4\% in mAP. $A^p$ refers to only adopting the pose alignment adjacency graph in the feature propagation module, we stack two feature propagation layers in our experiments. Compared with \textit{+Attention}, $+A^p$ improves Rank-1 and mAP accuracy by 0.7\% and 0.4\%. $A^f$ refers to utilizing feature affinity graph, similar to $A^p$, we stack two feature propagation layers, and we can obtain 88.5\% in Rank-1 and79.8\% in mAP. Furthermore, we can achieve 89.3\% and 80.4\% on MARS dataset by combining $A^p$ and $A^f$. With the consistency loss in the training stage, we can find the Rank-1 and mAP accuracy are improved by 0.5\% and 0.7\% respectively. With all these proposed components, we improve the Rank-1 and mAP accuracies from 87.8\% and 78.0\% to 89.8\% and 81.1\% respectively.

\begin{figure}[ht]
    \begin{center}
        \includegraphics[width=1\linewidth]{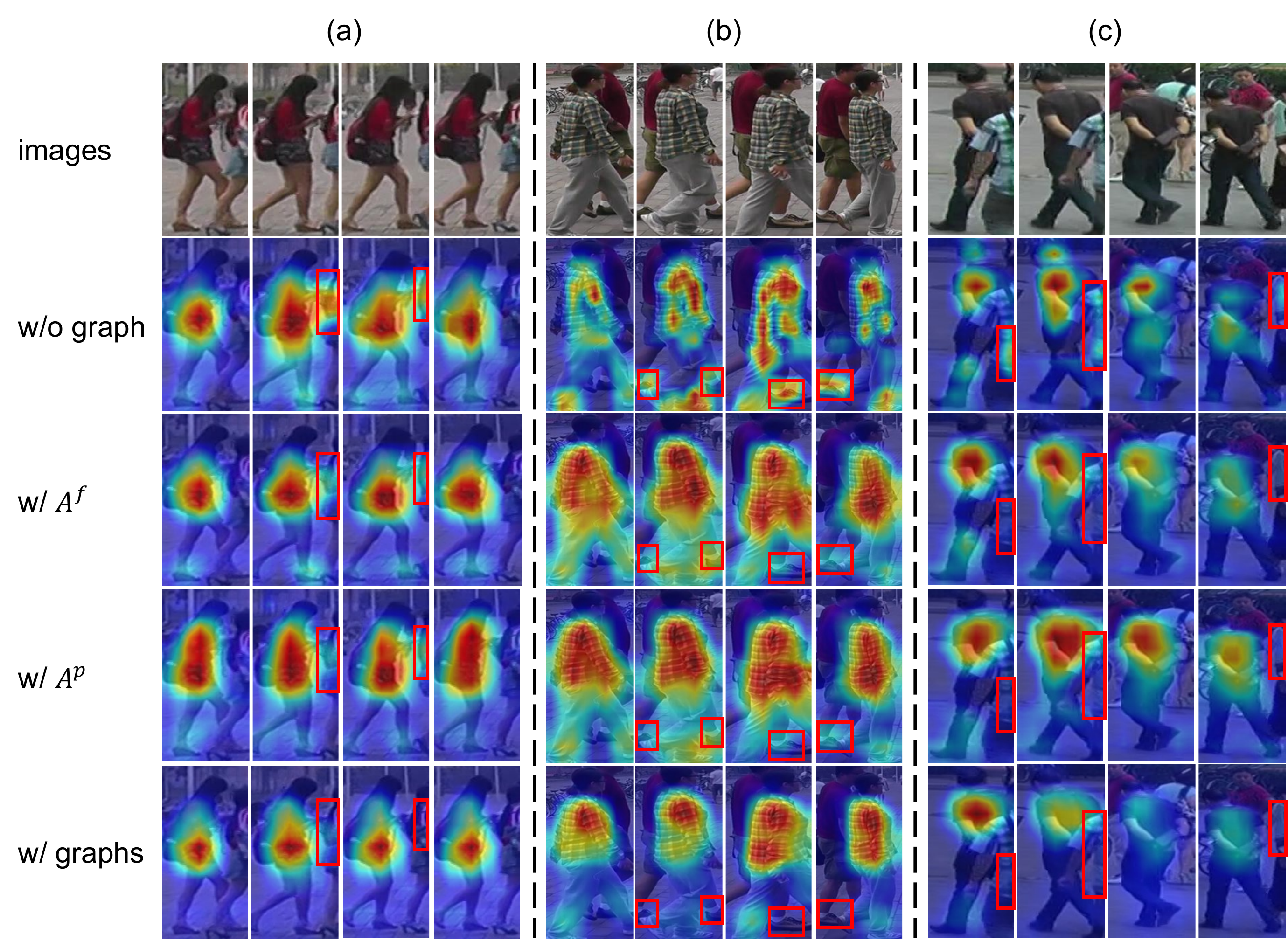}
    \end{center}
   \caption{Qualitative results for the baseline model (w/o graph) and our proposed method (w/ $A^f$, w/ $A^p$, and w/ graphs). We employ Grad-CAM to visualize the regions focused by the Re-ID model. The first row is the image sequences sampled from videos. In the other rows, the class activation maps for the baseline model and our proposed method are provided. Compared with the model without graph learning, we find out that our proposed method is robust to the occlusion and clutter background. \textbf{Best viewed in color and zoomed in.}}
\label{fig:visualization}
\end{figure}

\textbf{Qualitative results.}
We visualize the qualitative results in Figure~\ref{fig:visualization} with Grad-CAM~\cite{Selvaraju_2017_ICCV}, which is popularly used in computer vision problems for a visual explanation. We show the cases for clutter background in image sequence (a) and (b), when the distractions appear in the background, CAM of the model without graph highlights the distractions in red boxes, while for the models with graph, the influence of distractions is weakened. For occlusion cases shown in image sequence (c), the identity is partially occluded by the other person, the wrong attentive regions are highlighted in CAM for the model without the graph, and the models with the graph are not influenced by the occlusion. In general, our proposed graph-based methods are capable of alleviating the impact of clutter background and occlusion.

\begin{figure*}[ht]
    \begin{center}
        \includegraphics[width=0.9\linewidth]{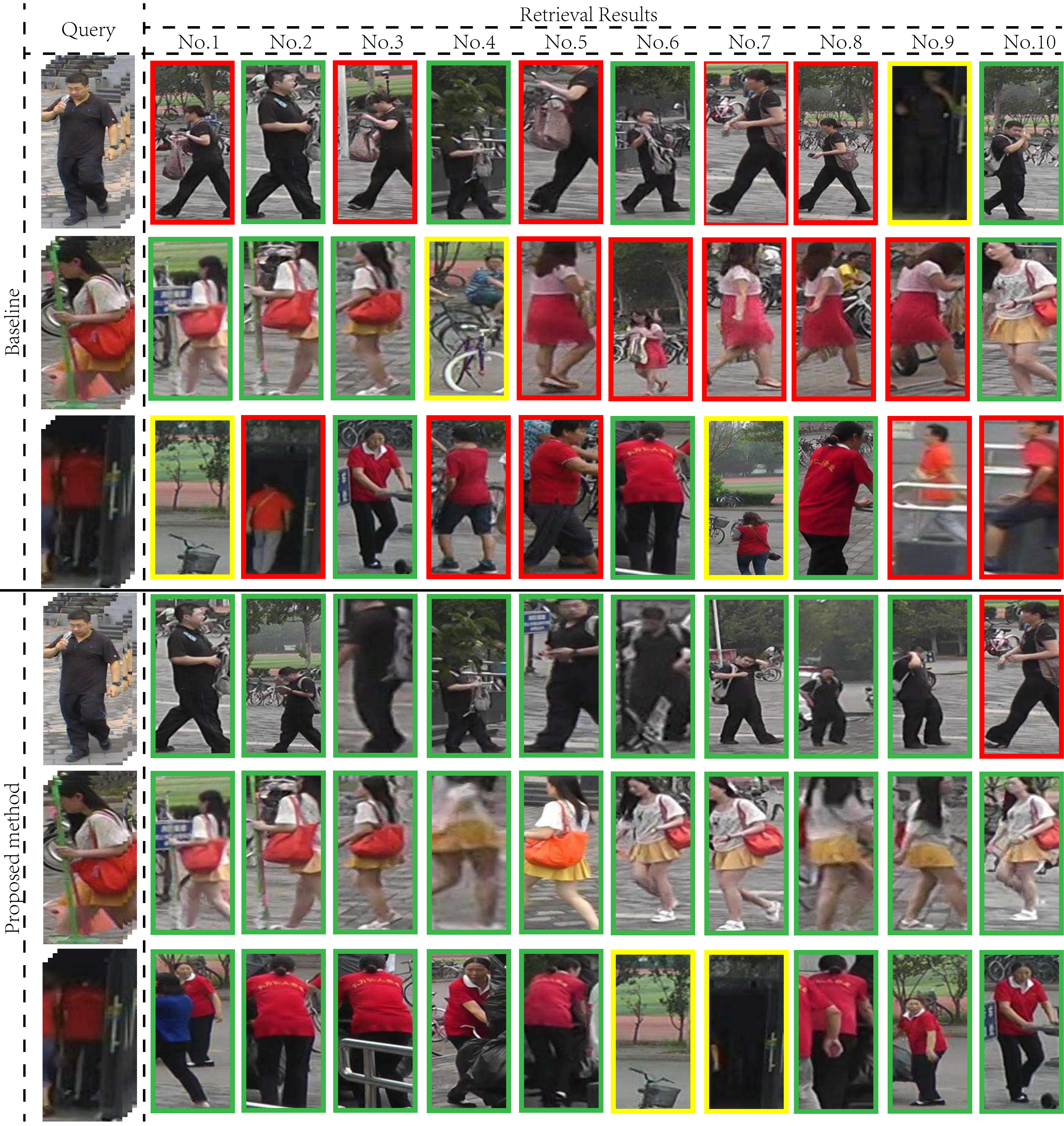}
    \end{center}
    \caption{Comparison of Rank-10 of our proposed method and baseline model. In each row, the images present the videos in the gallery. The images with the green box match the query, and the red box is the wrong matched result. Besides, the images with the yellow box are distracters, which is neglected while calculating the accuracy. \textbf{Best viewed in color.}}
    \label{fig:retrieval}
\end{figure*}

\textbf{Retrieval Results}
As illustrated in Figure~\ref{fig:retrieval}, we provide the retrieval results on MARS dataset for the baseline model and our proposed method. The illustration presents the improvement of our proposed method.

\section{Conclusion}\label{sec:conclusion}
This paper proposes an innovative graph representation learning approach for video person Re-ID. The proposed method can learn an adaptive structure-aware adjacency graph over the spatial person regions. By aggregating the contextual messages from neighbors for each node, the intrinsic affinity structure information among person feature nodes is captured adaptively, and the complementary contextual information is further propagated to enrich the person feature representations. Furthermore, we propose a novel regularization to enforce the consistency among different temporal resolutions, and it is beneficial for learning the compact and discriminative representations. The experimental results on four standard benchmarks demonstrate the effectiveness of the proposed scheme, and extensive ablation studies validate the feasibility of components in the network.

% \ifCLASSOPTIONcompsoc
%   % The Computer Society usually uses the plural form
%   \section*{Acknowledgments}
% \else
%   % regular IEEE prefers the singular form
% \fi

% \ifCLASSOPTIONcaptionsoff
%   \newpage
% \fi

\bibliographystyle{ieeetran}
\bibliography{wu}

\vfill

\end{document}